%% file: main.tex
\pdfoutput=1

\documentclass[11pt]{article}

\usepackage[]{acl}

\usepackage{times}
\usepackage{latexsym}

\usepackage[T1]{fontenc}

\usepackage[utf8]{inputenc}

\usepackage{microtype}

\usepackage{inconsolata}

\usepackage{amsmath}
\usepackage{adjustbox}
\usepackage{multirow}
\usepackage{graphicx}
\usepackage{subfig}
\usepackage{colortbl}

\title{Towards Robust Extractive Question Answering Models: Rethinking the Training Methodology}
\author{Son Quoc Tran$^{\S \thanks{This work was performed during undergraduate study at Denison University.}}$, Matt Kretchmar $^{\dagger}$\\
$^\S$Cornell University, Ithaca, NY\\
$^\dagger$Denison University, Granville, OH \\ 
\texttt{sontran@cs.cornell.edu, kretchmar@denison.edu}}

\begin{document}
\newcommand{\todo}[1]{\textcolor{red}{\textbf{To-do:} #1}}

\maketitle

\begin{abstract}
This paper proposes a novel training method to improve the robustness of Extractive Question Answering (EQA) models. Previous research has shown that existing models, when trained on EQA datasets that include unanswerable questions, demonstrate a significant lack of robustness against distribution shifts and adversarial attacks. Despite this, the inclusion of unanswerable questions in EQA training datasets is essential for ensuring real-world reliability. 
Our proposed training method includes a novel loss function for the EQA problem and challenges an implicit assumption present in numerous EQA datasets. Models trained with our method maintain in-domain performance while achieving a notable improvement on out-of-domain datasets. This results in an overall F1 score improvement of 5.7 across all testing sets. Furthermore, our models exhibit significantly enhanced robustness against two types of adversarial attacks, with a performance decrease of only about a third compared to the default models. \footnote{Our code and training data are publicly available at \href{https://github.com/sonqt/robust_qa}{github.com/sonqt/robust\_qa}.}
\end{abstract}
\input{sections/1-introduction}

\input{sections/2-related-work}
\input{sections/3-tasks-models}

\input{sections/4-attack}
\input{sections/5-objectives}
\input{sections/6-experiments}
\input{sections/7-discussion}
\input{sections/8-conclusion}

\bibliography{custom}

\clearpage

\input{sections/appendix}

\end{document}

%% file: sections/1-introduction.tex
\section{Introduction}
\label{sec:introduction}
Unanswerable questions are a valuable part in the training datasets of Extractive Question Answering (EQA) models. By learning from these questions, models can develop the ability to avoid extracting misleading responses, ultimately improving their reliability in real-world applications.   

Currently, there are two lines of research on unanswerable questions in EQA. Firstly, \citet{rajpurkar-etal-2018-know} introduced the SQuAD 2.0 dataset by adding  \textbf{\textit{adversarial unanswerable questions}} into SQuAD 1.1 \cite{rajpurkar-etal-2016-squad}. This work later inspired similar benchmarks in other languages such as French \cite{https://doi.org/10.48550/arxiv.2109.13209} and Vietnamese \cite{https://doi.org/10.48550/arxiv.2203.11400}. In the crowdsourcing process for adversarial unanswerable questions, human annotators are typically presented with a triple of context, an answerable question, and its corresponding answer(s). They are then asked to write unanswerable questions that exhibit an adversarial similarity to the presented answerable ones. 

In addition to the adversarially-written unanswerable questions, Natural Question \cite{kwiatkowski-etal-2019-natural}, Tydi QA \cite{clark-etal-2020-tydi}, and SQuAD \textit{AGent} \cite{tran-etal-2023-impacts} propose more naturally constructed unanswerable questions. This category of unanswerable questions is also known as \textbf{\textit{information-seeking unanswerable questions}}, emerging within the realm of information retrieval. These questions are initially independent of any context. The contexts are then paired with the questions as a result of the attempt to locate answers for the given questions within a large database containing multiple contexts.

The distinct characteristics of these two types of unanswerable questions pose a challenge for models. 
Models trained with one type of unanswerable questions often struggle when encountering the other type \cite{sulem-etal-2021-know-dont, tran2023agent}, defined in Machine Learning as a lack of robustness under distribution shift in the inputs. Additionally, models trained on unanswerable questions also demonstrate a lack of robustness against adversarial attack \cite{tran-etal-2023-impacts}. Notably, models trained on adversarial unanswerable questions in SQuAD 2.0 tend to output an ``empty'' response upon detecting any sign of contradiction between the attack sentence and the given question.

We hypothesize that the observed lack of robustness in EQA models can be attributed to two primary factors.
First, the current EQA training loss objective \cite{devlin-etal-2019-bert} inaccurately treats unanswerable questions as if they have an answer span. This span is designated to start and end at the special classification token \texttt{[CLS]} of the pre-trained model which is also the first token in the input sequence. This approach potentially misguides the model's understanding of unanswerable questions. Second, the assumption that a given question can only have a single answer or no answer introduces a learning shortcut, making EQA models vulnerable to adversarial attacks.

In this work, we propose a new training method for EQA models to address the two problems discussed above. First, we design new training loss function that naturally treats unanswerable questions as lacking any answer. Second, to overcome the single-answer assumption in most EQA datasets, we create a new ``synthetic'' answer span in a number of answerable questions. Our empirical findings are summarized as follows:
\begin{enumerate}
    \item We test our newly proposed training method on three language models. While the new method does not reduce the in-domain performance of models, models fine-tuned with our training method show a $13$ F1-score improvement on out-of-domain testing sets. Furthermore, our models exhibit significantly enhanced robustness against two types of adversarial attacks, with a performance decrease of only $13.2$ in F1-score compared to a $40.7$ decrease of default models.
    \item We also investigate the independent contributions of new loss function and ``synthetic'' answers in our training method. Our analysis reveals that the new loss function helps enhance the robustness against distribution shifts from adversarial unanswerable questions in the training set to information-seeking unanswerable questions in the testing set. On the other hand, eliminating the single-answer assumption by creating ``synthetic'' answer significantly enhances the robustness of models against adversarial attacks.
\end{enumerate}

%% file: sections/2-related-work.tex
\section{Related Work}
There are two key research areas on improving the robustness of natural language processing (NLP) models: robustness against adversarial attacks and against distribution shift \cite{wang-etal-2022-measure}. Adversarial attacks  involve editing a test sample to create a more challenging example for trained models without causing additional difficulty for humans. These attacks can be classified based on whether the attack process has access to the models' parameters (white-box attacks, \cite{blohm-etal-2018-comparing,neekhara-etal-2019-adversarial,alzantot-etal-2018-generating,wallace-etal-2019-universal,ebrahimi-etal-2018-hotflip}) or not (black-box attacks, \cite{jia-liang-2017-adversarial,ribeiro-etal-2018-semantically,wang-bansal-2018-robust,blohm-etal-2018-comparing, iyyer-etal-2018-adversarial}). On the other hand, robustness against distribution shift is measured using test samples that exhibit linguistic differences from the samples encountered by models during the training phase \cite{pmlr-v119-miller20a}.

Findings of limited robustness in NLP models have spurred significant efforts to improve their resilience. From a data-driven perspective, adversarial attacks can be employed during the training phase to enhance model robustness. Augmented training data can be created by heuristically editing \cite{wang-bansal-2018-robust} or through neural-based generation \cite{iyyer-etal-2018-adversarial, khashabi-etal-2020-bang, bartolo-etal-2021-improving, fu-etal-2023-scene}. Additionally, increasing the diversity of training data has proven to be an effective strategy for improving model robustness \cite{fisch-etal-2019-mrqa, khashabi-etal-2020-unifiedqa}.

In addition to data-driven approaches, model-based approaches are also effective in improving model robustness. Following the success of BERT, various studies have shown that the pretraining process, which involves a self-supervised objective and the use of large amounts of diverse pretraining data, significantly enhances the generalization of language models in downstream tasks \cite{hendrycks-etal-2020-pretrained, tu-etal-2020-empirical}.

Another research direction involves using a biased model during the training phase to force the target model to discard some spurious patterns in the training set. These biased models can be designed with a specific targeted type of bias \cite{clark-etal-2019-dont, schuster-etal-2019-towards, he-etal-2019-unlearn, utama-etal-2020-mind, karimi-mahabadi-etal-2020-end}, or without prior knowledge about the biases present in the training dataset \cite{clark-etal-2020-learning, utama-etal-2020-towards, ghaddar-etal-2021-end, sanh2021learning}.

Our work distinguishes itself by combining both data-driven and model-driven approaches. 
From data-driven side, we challenge the implicit assumption of single answers in multiple current EQA datasets by augmenting ``synthetic'' answers to a number of training samples. We hope that our experimental results with synthetic answers will inspire the development of EQA datasets that incorporate multi-span questions, enabling answers composed of multiple non-adjacent spans of text \cite{li-etal-2022-multispanqa}.
On the model-driven side, we propose a novel training loss for EQA models that enhances their robustness against distribution shift of unanswerable questions.
With these novel approaches, we aim to extensively improve the robustness of models against both distribution shifts and adversarial attacks.

%% file: sections/3-tasks-models.tex
\section{Models and Tasks}
In Extractive Question Answering (EQA), models are trained to identify the answer (a text span in the context) to the given question. The dataset may include unanswerable questions, for which a valid prediction is an ``empty'' answer. A common metric to evaluate MRC systems is F1-score. It measures the average overlap between the words in the predicted answer and the human-annotated gold answer.
\subsection{Models}
\input{tables/attack-example}

In this work, we evaluate our newly proposed training method using the base version of three pre-trained models BERT \cite{devlin-etal-2019-bert}, RoBERTa \cite{liu2019roberta}, and SpanBERT \cite{joshi-etal-2020-spanbert}.
\subsection{Extractive Question Answering}
An EQA problem is given by  a test set $\mathcal{D}$ of triplets $(q,c,a)$ where $q$ is a question posed to models, $c$ is the corresponding context (usually a short paragraph of text), and $a$ is the expected answer (or set of ``gold'' answers). The performance of the EQA model $f$ is measured by 
\begin{eqnarray*}
Per(f, \mathcal{D}) = \frac{1}{\mid \mathcal{D} \mid} \sum_{(c,q,a) \in \mathcal{D}} m(a, f(c,q))
\end{eqnarray*}
where $m$, in this paper, is the F1-score metric.

In our experiments, we evaluate models on both answerable and unanswerable questions from different domains as outlined in the next section. To compare the performance of models across all tested domains, we assume that (1) the number of answerable questions is equal to the number of unanswerable questions, and that (2) the importance of different domains is the same.
\begin{align*}
Per(f) = \frac{Per_{has-ans}(f) + Per_{no-ans}(f)}{2}
\end{align*}

where $Per_{has-ans}(f)$ and $Per_{no-ans}(f)$ are the average performance of model $f$ on all domains of answerable and unanswerable questions, respectively. Specifically, we can calculate $Per_{has-ans}(f)$ as follows: 
\begin{align*}
Per_{has-ans}&(f) =\\& \frac{1}{\mid \mathcal{S}^{has-ans} \mid} \sum_{\mathcal{D} \in \mathcal{S}^{has-ans}} Per(f, \mathcal{D})
\end{align*}
, where $\mathcal{S}^{has-ans}$ is the set of all testing set with answerable questions.
\subsection{Datasets}
In our experiments, we fine-tune our EQA models by conducting additional training on SQuAD 2.0 \cite{rajpurkar-etal-2018-know} (for Sections \ref{sec:experiment} or \ref{sec:discussion}) and SQuAD \textit{AGent} \cite{tran2023agent} (for Section \ref{sec:discussion}). While both datasets share the same answerable questions, SQuAD 2.0 includes adversarially written unanswerable questions, whereas SQuAD \textit{AGent} utilizes information-seeking unanswerable questions.

We test the performance of our models on

\begin{itemize}
\item \textbf{SQuAD 2.0}: We test our models on both \textit{\textbf{answerable}} (\textit{has-ans}) and \textit{\textbf{unanswerable}} (\textit{no-ans}) questions of this dataset. The unanswerable questions in SQuAD 2.0 are adversarially written.
\item \textbf{SQuAD \textit{AGent}}: We only test models on \textbf{\textit{unanswerable}} questions (\textit{AGent}) of this dataset because the answerable questions in this dataset are the same as ones in SQuAD 2.0. The unanswerable questions from this dataset are information-seeking.
\item \textbf{ACE-whQA} \cite{sulem-etal-2021-know-dont}: We test models on \textbf{\textit{answerable}} (\textit{has-ans}) questions and \textbf{\textit{two types of unanswerable}} questions: competitive (\textit{no-ans competitive}), where the passage contains an entity of the same type as the expected answer, and non-competitive (\textit{no-ans non-com}), where the passage does not contain any entity of the same type as the expected answer.

\end{itemize}
The diversity of testing domains enables us to measure the robustness of models against distribution shifts, which occur when encountering testing data that differs from the training data.

%% file: tables/attack-example.tex
\begin{table*}[ht]
\centering
\resizebox{15cm}{!}{%
\begin{tabular}{clll}
\hline
Attack Types & Question & Attacked Context & \begin{tabular}[c]{@{}c@{}}Ground Truth \\ Answer \end{tabular}\\ \hline
\begin{tabular}[c]{@{}c@{}}AddOneSent \\ \textit{AOS} \\ \cite{jia-liang-2017-adversarial}\end{tabular} & \begin{tabular}[c]{@{}l@{}}What is the name\\  of the water body \\ that is found to\\ the east?\end{tabular} & \begin{tabular}[c]{@{}l@{}}To the east is the Colorado Desert and the \\ \textcolor{blue}{\textbf{Colorado River}} at the border with Arizona, \\ and the Mojave Desert at the border with the \\ state of Nevada. To the south is the Mexico\\–United States border. \textcolor{red}{\textbf{Sea is the name of}} \\ \textcolor{red}{\textbf{the water body that is found to the west.}}\end{tabular} & \textcolor{blue}{\textbf{Colorado River}} \\ \hline
\begin{tabular}[c]{@{}c@{}}Negation\\ \cite{tran-etal-2023-impacts} \end{tabular}& \begin{tabular}[c]{@{}l@{}}What is the name\\  of the water body \\ that is found to\\ the east?\end{tabular} & \begin{tabular}[c]{@{}l@{}}To the east is the Colorado Desert and the \\ \textcolor{blue}{\textbf{Colorado River}} at the border with Arizona, \\ and the Mojave Desert at the border with the \\ state of Nevada. To the south is the Mexico\\–United States border. \textcolor{red}{\textbf{Sea is the name of}} \\ \textcolor{red}{\textbf{the water body that is found to the not east.}}\end{tabular} & \textcolor{blue}{\textbf{Colorado River}} \\ \hline
\end{tabular}
}
\caption{Examples of AddOneSent (\textit{AOS}) and Negation Attacks on answerable questions. The adversarial sentence is highlighted in red color.}
\label{tab:attack-example}
\end{table*}

%% file: sections/4-attack.tex
\section{Adversarial Attacks}
In addition to evaluating models' robustness against distribution shift, we also measure the robustness against adversarial attacks.
\subsection{Robustness Evaluation}
\label{sec:robustness-eval}

An attack algorithm $\mathcal{A}$ transforms triplets $(q,c,a)$ in $\mathcal{D}$ into adversarial test samples $(q',c',a')$ in the adversarial test set $\mathcal{D}^{\mathcal{A}}_{attacked}$,
where $c'$, $q'$, and $a'$ are the modified (attacked) versions of $c$, $q$, and $a$. The robustness of a model is then computed as the difference between the performance of the model on the original test set vs attacked test set: 
\begin{eqnarray*}
\Delta^{\mathcal{A}} = Per(f, \mathcal{D}) - Per(f, \mathcal{D}^{\mathcal{A}}_{attacked})
\end{eqnarray*}

When there are more than one attack algorithm, we measure the overall robustness by 
\begin{eqnarray*}
\Delta = \frac{1}{\mid \mathcal{T} \mid}\sum_{\mathcal{A} \in \mathcal{T}} \Delta^{\mathcal{A}}
\end{eqnarray*}
where $\mathcal{T}$ is the set of all tested types of adversarial attacks. 

\subsection{Algorithms for Attack Construction}
In this paper, we test the experimented models on two types of adversarial attacks. 
\subsubsection{AddOneSent Attacks}
Table \ref{tab:attack-example} gives an example of AddOneSent (\textit{AOS}) attack \cite{jia-liang-2017-adversarial}. 
The \textit{AddOneSent attack} strategy creates the attack sentence from a modified question and a fake answer. To construct the modified question, nouns and adjectives in the original question are substituted with their antonyms sourced from WordNet \cite{10.7551/mitpress/7287.001.0001}. Meanwhile, the fake answer is nearest word to the original gold answer in the vector space of GloVe \cite{pennington-etal-2014-glove}.

\subsubsection{Negation Attacks}

The Negation Attack, shown in Table \ref{tab:attack-example}, is designed to mislead models into giving incorrect ``empty'' predictions. This method involves the crafting of an attack statement that has significant lexical overlap with the original question yet is easy to identify as contradictory by simply inserting ``not'' in front of the first adjective within the question. The fake answer is created similarly to the AddOneSent attack.

The questions and answers are unchanged in both types of attacks ($q'=q$ and $a'=a$). 

%% file: sections/5-objectives.tex
\section{Extractive Question Answering Loss Functions}

EQA models are typically fed a question  $q$ and a context $c$ as input. State-of-the-art EQA models, often employing BERT-style language models at their core, process  $q$ and $c$ together as a sequence input $<\texttt{[CLS]}q\texttt{[SEP]}c>$, with \texttt{[CLS]} and \texttt{[SEP]} as special tokens of pre-trained tokenizer accompanying the pre-trained model.

Given an input sequence (pair of question-context) with $n$ tokens $seq = (t_1, t_2, ..., t_n)$, we have $$\mathcal{M}(seq) = (\vec{v_1}, \vec{v_2}, ..., \vec{v_n})$$
where $\mathcal{M}$ is a pre-trained language model that takes sequence $seq$ as the input and output $n$ contextualized vectors $(\vec{v_1}, \vec{v_2}, ..., \vec{v_n})$, each corresponds to one of the input tokens, encoding its contextual information.

We then employ two single-layer feed-forward neural networks, denoted as $S$ and $E$ for predicting the start and end positions, respectively. Both networks are designed to receive input vectors $\vec{v_k}$ and produce a scalar output. We then have that
$$s_k = S(\vec{v_k}),\quad e_k = E(\vec{v_k})$$
for every $\vec{v_k}$ in $(\vec{v_1}, \vec{v_2}, ..., \vec{v_n})$.

\subsection{Default Loss Function}
\citet{devlin-etal-2019-bert} use the Cross Entropy loss function for training BERT on SQuAD 2.0.
\begin{align*}
L_{Default} &= -\Sigma_{k=1}^n \log \frac{\exp(s_k)}{\Sigma_{i=1}^n \exp(s_i)}y^s_k \\
         & \qquad -\Sigma_{k=1}^n \log \frac{\exp(e_k)}{\Sigma_{i=1}^n \exp(e_i)}y^e_k
\end{align*}
where $y^s_k$ and $y^e_k$ are the labels of whether $k^{th}$ token in the input sequence is the start or end of a gold answer identified by human annotators. Unanswerable questions are treated as having an answer span with start and end at the \texttt{[CLS]} token, which means $y^s_0$ and $y^e_0$ are $1$s.

As of the time of writing this paper, the training methodology utilizing this particular loss function remains widely adopted in most EQA models. We term this training methodology the ``default'' approach.

\subsection{Our Loss Function}
\subsection*{QA Loss}
This component ($L_{QA}$) of the newly proposed loss function is similar to the Cross Entropy loss function used in work by \citet{devlin-etal-2019-bert}.
However, a key difference lies in how we handle unanswerable sequences. In our approach, since we treat all tokens in these sequences as equally unlikely to be the start or end of an answer, all tokens within an unanswerable sequence are assigned the same label uniformly,  represented as $y_k^s = y_k^e = \frac{1}{n}$, where $n$ denotes the sequence length.

Note that setting these all labels to 0 would result in no backpropagation signal for unanswerable sequences. By using a ground truth of $\frac{1}{n}$ for $n$ tokens, the sum of these values equals $1$, which is an appropriate scale for the output of the softmax function of the Cross Entropy loss.
\subsection*{Sequence Tagging Loss}
We enable our models to naturally signal ``unanswerable'' predictions by using an inference pipeline that outputs an ``empty'' prediction if the maximum span score of $s_i + e_j$ is negative. To enable models to output negative $s_i + e_j$ scores for all spans in unanswerable sequences, we incorporate sequence tagging loss alongside the standard QA loss:
\begin{align*}
&L_{Tag} = \\
&- \Sigma_{k=1}^n (y^s_k  \log \sigma(s_k)+(1-y^s_k) \log(1-\sigma(s_k))) \\
&- \Sigma_{k=1}^n (y^e_k  \log \sigma(e_k) +(1-y^e_k) \log(1-\sigma(e_k)))
\end{align*}
where $\sigma(x) = \frac{1}{1+\exp(-x)}$, the labels for the gold start tokens are assigned $y^s_k = 1$, and labels for all other tokens are set to $y^s_k = 0$. This logic extends to the labels for end tokens. Consequently, all $y^s_k$ and $y^e_k$ in unanswerable sequences are zeros.

\subsection*{Overall Loss}
$$L_{Ours} = \lambda_{QA}\cdot L_{QA} + \lambda_{Tag}\cdot L_{Tag}$$
where $\lambda_{QA}$ and $\lambda_{Tag}$ denote weights for their corresponding losses. In this paper, we set $\lambda_{QA}= 2$ and $\lambda_{Tag} = 1$. Appendix \ref{appendix:lamba-dis} discusses the selection of these weights in more detail.

\input{tables/main-ood}

\subsection{Inference Pipeline}
In both model types, the score for a candidate span ranging from position $i$ to position $j$ is given by $s_i + e_j$, The span with the highest score, where $j\geq i$, is selected for prediction.

Models trained with the default training loss function indicate an unanswerable question by outputting an ``empty'' string when the highest scoring span is $(0,0)$, which corresponds to the \texttt{[CLS]} token.

Conversely, models trained with our method indicate an ``empty'' string response when the maximum span score of $s_i + e_j$ is negative.

%% file: tables/main-ood.tex
\begin{table*}[ht]
\centering
\resizebox{15cm}{!}{%
\begin{tabular}{lc|ccc|ccc|cc|c}
\hline
\multicolumn{2}{c|}{\multirow{2}{*}{\begin{tabular}[c]{@{}c@{}}\textbf{\textit{Train Set:}} \\ \textbf{\textit{SQuAD 2.0}} \end{tabular}}} & \multicolumn{3}{c|}{\textbf{SQuAD}} & \multicolumn{3}{c|}{\textbf{ACE-whQA}} & \multicolumn{2}{c|}{\textbf{Average}} & \multirow{2}{*}{\textbf{Overall}} \\ \cline{3-10}
 &  & \cellcolor[HTML]{C0C0C0}has-ans & \cellcolor[HTML]{C0C0C0}no-ans & \textit{AGent} & has-ans & \begin{tabular}[c]{@{}c@{}}no-ans\\ non-com\end{tabular} & \begin{tabular}[c]{@{}c@{}}no-ans\\ competitive\end{tabular} & has-ans & no-ans &  \\ \hline
\multirow{2}{*}{\textbf{BERT}} & Default & \cellcolor[HTML]{C0C0C0}\textbf{78.8} & \cellcolor[HTML]{C0C0C0}71.1 & 44.2 & 67.6 & 52.3 & \textbf{38.7} & \textbf{73.2} & 51.6 & 62.4 \\
 & \textit{\textbf{Ours}} & \cellcolor[HTML]{C0C0C0}73.7 & \cellcolor[HTML]{C0C0C0}\textbf{75.7} & \textbf{63.2} & \textbf{69.9} & \textbf{59.1} & 36.6 & 71.8 & \textbf{58.7} & \textbf{65.3} \\ \hline
\multirow{2}{*}{\textbf{RoBERTa}} & Default & \cellcolor[HTML]{C0C0C0}\textbf{85.0} & \cellcolor[HTML]{C0C0C0}81.2 & 51.8 & 66.0 & 77.1 & 57.8 & \textbf{75.5} & 67.0 & 71.3 \\
 & \textit{\textbf{Ours}} & \cellcolor[HTML]{C0C0C0}81.3 & \cellcolor[HTML]{C0C0C0}\textbf{85.6} & \textbf{67.9} & \textbf{67.4} & \textbf{85.3} & \textbf{66.3} & 74.4 & \textbf{76.3} & \textbf{75.4} \\ \hline
\multirow{2}{*}{\textbf{SpanBERT}} & Default & \cellcolor[HTML]{C0C0C0}\textbf{86.0} & \cellcolor[HTML]{C0C0C0}76.0 & 46.0 & \textbf{66.0} & 53.1 & 24.2 & \textbf{76.0} & 49.8 & 62.9 \\
 & \textit{\textbf{Ours}} & \cellcolor[HTML]{C0C0C0}80.2 & \cellcolor[HTML]{C0C0C0}\textbf{81.9} & \textbf{66.1} & 61.5 & \textbf{90.5} & \textbf{60.4} & 70.9 & \textbf{74.7} & \textbf{72.8} \\ \hline
\multirow{2}{*}{\textbf{Average}} & Default & \cellcolor[HTML]{C0C0C0}\textbf{83.3} & \cellcolor[HTML]{C0C0C0}76.1 & 47.3 & \textbf{66.5} & 60.8 & 40.2 & \textbf{74.9} & 56.1 & 65.5 \\
 & \textit{\textbf{Ours}} & \cellcolor[HTML]{C0C0C0}78.4 & \cellcolor[HTML]{C0C0C0}\textbf{81.1} & \textbf{65.7} & 66.3 & \textbf{78.3} & \textbf{54.4} & 72.4 & \textbf{69.9} & \textbf{71.2} \\ \hline
\end{tabular}
}
\caption{Performance of models fine-tuned on SQuAD 2.0 using Default training method and our proposed training method, each averaged over five runs with random initialization. The performance on in-domain samples are highlighted in gray cells.}
\label{tab:main-ood}
\end{table*}

%% file: sections/6-experiments.tex
\section{Experiments}
\label{sec:experiment}
\subsection{Experiment Design}
\label{sec:ex-design}
In the experiments in this section, we train our models using the SQuAD 2.0 dataset. For models trained with the default loss function, the original SQuAD 2.0 dataset is used without modifications. However, for models trained using our proposed method in this section, we introduce modifications to the SQuAD 2.0 dataset to eliminate the single-answer assumption during the training phase. We augment approximately $20\%$ of the answerable questions in the original dataset with an additional ``synthetic'' answer, resulting in these questions having two answers. In Appendix \ref{appendix:synthetic}, we provide a detailed information on how we generate "synthetic" answers, along with our experiments on the risks of hallucinations when training EQA models using these synthetic answers.

\subsection{Results}

Table \ref{tab:main-ood} shows performances of models trained on default and our training methods. Firstly, models trained with our method (new loss function and additional synthetic answers) achieve almost the same performance as those trained using default approach on SQuAD 2.0, the in-domain testing set. Specifically, models trained with the default loss function achieve an average F1 score of $79.7$ (across both answerable and unanswerable questions $\frac{83.3 + 76.1}{2}$) on SQuAD 2.0, while our models achieve an average F1 score of $79.8$.

\input{figures/progress}

On the other hand, our models consistently outperform default model on out-of-domain unanswerable questions, including those from SQuAD \textit{AGent} and both competitive and noncompetitive unanswerable questions from ACE-whQA. On information-seeking unanswerable questions from SQuAD \textit{AGent}, our models outperform default models by a large margin of $18.4$ F1 score on average. Furthermore, on the unanswerable questions in ACE-whQA, our models outperform default ones by $17.5$ F1 for noncompetitive unanswerable questions and $14.2$ F1 for competitive ones. This enhanced robustness against distribution shifts enables our models to attain a higher overall performance of $71.2$, compared to the $65.5$ achieved by default models across all evaluated answerable and unanswerable questions.

We then analyze the performance gap of each model on unanswerable questions between SQuAD 2.0 and SQuAD \textit{AGent} over three training epochs. Figure \ref{fig:progress} presents the dynamics of this performance gap for RoBERTa models trained with the default method and our proposed method on SQuAD 2.0. 

Notably, models using the default loss function exhibit an increasing performance gap throughout the training process. This indicates that as models better perform on adversarial unanswerable questions within SQuAD 2.0, their performance on information-seeking unanswerable questions in SQuAD \textit{AGent} decreases significantly. Conversely, models trained with our proposed loss function demonstrate a stable robustness against such shifts across three training epochs.

\input{tables/main-robustness}

In addition to evaluating the generalization of our models, we also evaluate their robustness against adversarial attacks. The results, presented in Table \ref{tab:main-attack-performance}, demonstrate the improved robustness of models trained with our method compared to those trained with the default approach. Specifically, under the AddOneSent attacks, the performance of default models drops by $27.4$, whereas our models exhibit a much smaller decrease of $9.9$ F1 score. Similarly, for the Negation attack, while default models experience a performance decrease of $56.3$, our models see a reduction of only $16.4$ on F1. These results highlight the significantly improved robustness of our models, with our training method mitigating $67.6\%$ of the performance drop due to adversarial attacks, reducing from $40.7$ to $13.2$ on F1-score metric.

%% file: figures/progress.tex
\begin{figure*}%
    \centering
    \subfloat[\centering Default]{{\includegraphics[width=0.46\textwidth]{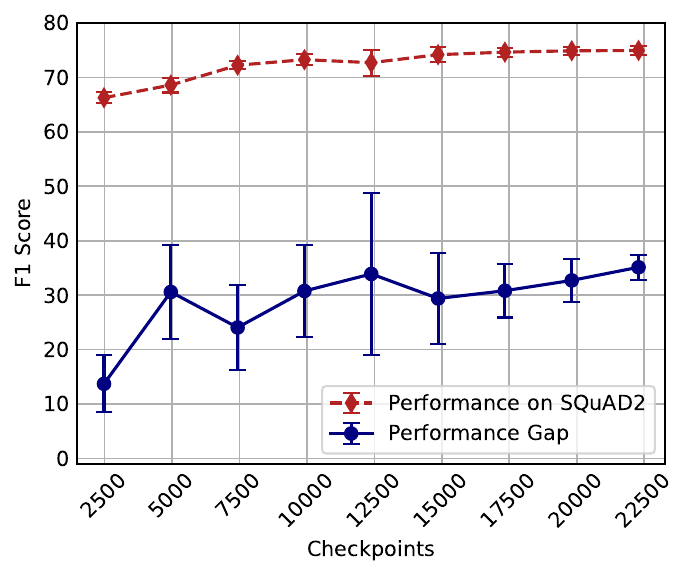} }}%
    \qquad
    \subfloat[\centering Ours ]{{\includegraphics[width=0.46\textwidth]{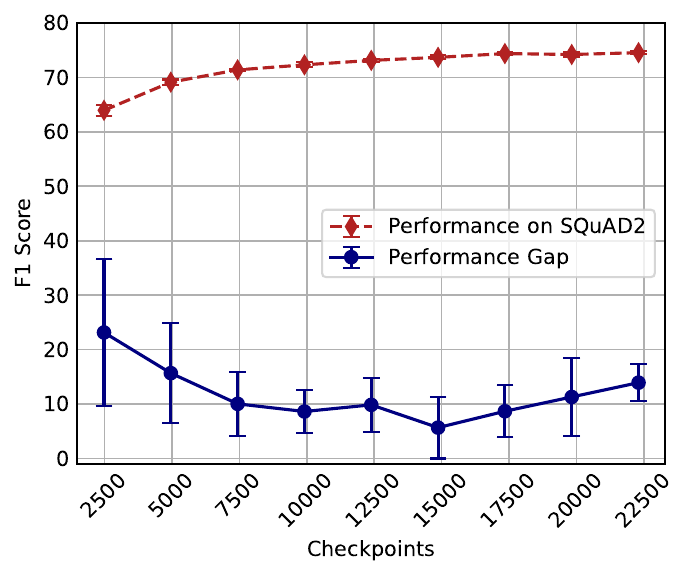} }}%
    \caption{The training dynamics of RoBERTa models trained using the Devlin method versus our proposed method on SQuAD 2.0. We analyze the performance gap on unanswerable questions between SQuAD 2.0 and SQuAD \textit{AGent} across three training epochs. The error bars represent the standard deviations of five runs.}

    \label{fig:progress}%
\end{figure*}

%% file: tables/main-robustness.tex
\begin{table}[ht]
\centering
\resizebox{\linewidth}{!}{
\begin{tabular}{lc|c|cc|c}
\hline
\multicolumn{2}{c|}{\multirow{2}{*}{\begin{tabular}[c]{@{}c@{}}\textbf{\textit{Train Set:}} \\ \textbf{\textit{SQuAD 2.0}} \end{tabular}}} & \multicolumn{1}{l|}{\multirow{2}{*}{Original}} & \multicolumn{2}{c|}{\textbf{\begin{tabular}[c]{@{}c@{}}Adversarial \\ Attack\end{tabular}}} & \multicolumn{1}{l}{\multirow{2}{*}{$\Delta \downarrow$}} \\ \cline{4-5}
\multicolumn{2}{l|}{} & \multicolumn{1}{l|}{} & \textit{AOS} & Negation & \multicolumn{1}{l}{} \\ \hline 
\multirow{2}{*}{\textbf{BERT}} & Default & \textbf{78.8} & 52.2 & 27.5 & 38.9 \\
 & \textit{\textbf{Ours}} & 73.7 & \textbf{64.0} & \textbf{49.5} & \textbf{16.9} \\ \hline
\multirow{2}{*}{\textbf{RoBERTa}} & Default & \textbf{85.0} & 56.1 & 30.9 & 41.5 \\
 & \textit{\textbf{Ours}} & 81.3 & \textbf{71.9} & \textbf{65.8} & \textbf{12.4} \\ \hline
\multirow{2}{*}{\textbf{SpanBERT}} & Default & \textbf{86.0} & 57.9 & 30.7 & 41.7 \\
 & \textit{\textbf{Ours}} & 80.2 & \textbf{69.5} & \textbf{70.6} & \textbf{10.1} \\ \hline
\multirow{2}{*}{\textit{\textbf{Average}}} & Default & \textbf{83.3} & 55.4 & 29.7 & 40.7 \\
 & \textit{\textbf{Ours}} & 78.4 & \textbf{68.5} & \textbf{62.0} & \textbf{13.2} \\ \hline
\end{tabular}
}
\caption{Robustness against adversarial attacks of models fine-tuned on SQuAD 2.0 using Default training method and our proposed training method.}
\label{tab:main-attack-performance}
\end{table}

%% file: sections/7-discussion.tex
\section{Further Analysis}
\label{sec:discussion}
\subsection{Experiment Design}
To evaluate the effectiveness of our proposed training method under different scenarios, we design two experiments.
\begin{enumerate}
    \item We train models on SQuAD 2.0 using our proposed loss function without introducing ``synthetic'' answers. We then compare these models (referred to as ``\textit{no synthetic}'') with those trained using the default loss function, also trained on SQuAD 2.0. This experiment is designed to study the independent contributions of the newly proposed loss function and the augmented ``synthetic'' answers to the robustness of our models.
    \item We train models on the information-seeking, unanswerable question dataset SQuAD \textit{AGent} using our proposed training method (including new loss function and ``synthetic'' answers). We then compare these models with those trained using the default method, also trained on SQuAD \textit{AGent}. This experiment investigates the effectiveness of our proposed method on datasets with information-seeking unanswerable questions.

\end{enumerate}
\subsection{Robustness against Distribution Shift}

\input{tables/squad2-nomultians-ood}

We now evaluate the performance of models trained on SQuAD 2.0 using our proposed loss function, while excluding synthetic answers. The experimental results, in Table \ref{tab:squad2-nomultians-ood}, highlight that even in the absence of synthetic answers, our models better generalize to information-seeking unanswerable questions. The ``\textit{No synthetic}'' outperform default models by a large margin of $18.4$ on F1 when tested on \textit{AGent} unanswerable questions. This finding shows that the robustness of our models can be mainly attributed to the incorporation of the new loss function.

Having established the successful generalization of our models from adversarial to information-seeking unanswerable questions, we now investigate the effectiveness of our loss function in achieving the reverse (generalizing from SQuAD \textit{AGent} to SQuAD 2.0).

\input{tables/agent-ood}

Table \ref{tab:agent-ood} shows the performance of models trained on SQuAD \textit{AGent} using default and our training methods. We observe that models trained with our method do not exhibit improved robustness against distribution shift to unanswerable questions in SQuAD 2.0, compared to those trained with the default method. This result indicates that our loss function mainly benefits the generalization of models to information-seeking unanswerable questions, such as those in SQuAD \textit{AGent}.

\subsection{Robustness against Adversarial Attacks}
\input{tables/agent-robustness}
While models trained with our method on SQuAD \textit{AGent} do not exhibit improved robustness against distribution shifts to SQuAD 2.0, they demonstrate significant improveme
nts when encountering adversarial attacks.

The experimental results in Table \ref{tab:agent-attack-performance} show that when using SQuAD \textit{AGent} as the training set, models trained with default approach exhibit a significant reduction in performance of $30.0$ F1 points. Conversely, models trained with our method (new loss function and the synthetic answers) experience a much smaller performance drop of $16.2$ F1 points. Our findings conclusively demonstrate that our training method notably enhances the robustness of models trained on both SQuAD 2.0 and SQuAD \textit{AGent} against adversarial attacks.

\input{tables/squad2-nomultians-robustness}

With this significant improvement established, we then shift our focus to identifying the primary factor behind this increased robustness. We hypothesize that our models' robustness against adversarial attacks might be mainly thanks to the augmented ``synthetic'' answers, which eliminate the single-answer assumption in the SQuAD dataset.

Therefore, we examine the robustness against adversarial attacks of ``\textit{no synthetic}'' models trained on SQuAD 2.0 using our proposed loss function, while omitting synthetic answers. The experimental results, in Table \ref{tab:squad2-nomultians-attack-performance}, indicate that without the synthetic answers, our models are no longer robust against adversarial attacks. The performance gap $\Delta$ of our models without synthetic answers is even higher than that of default models ($41.3$ compared to $40.7$). This finding strongly supports our hypothesis that the inclusion of ``synthetic'' answers in our training method is a key factor in the improved robustness against adversarial attacks of our models.

In Appendix \ref{appendix:synthetic},  we further validate this hypothesis by training models on SQuAD 1.1 \cite{rajpurkar-etal-2016-squad}, a dataset that contains only answerable questions.

%% file: tables/squad2-nomultians-ood.tex
\begin{table}[ht]
\centering
\resizebox{\linewidth}{!}{
\begin{tabular}{lc|ccc}
\hline
\multicolumn{2}{c|}{\multirow{2}{*}{\begin{tabular}[c]{@{}c@{}}\textbf{\textit{Train Set:}} \\ \textbf{\textit{SQuAD 2.0}} \end{tabular}}}  & \multicolumn{3}{c}{\textbf{SQuAD}} \\ 
 &  & \cellcolor[HTML]{C0C0C0}has-ans & \cellcolor[HTML]{C0C0C0}no-ans & \textit{AGent} \\ \hline
\multirow{2}{*}{\textbf{BERT}} & Default & \cellcolor[HTML]{C0C0C0}\textbf{78.8} & \cellcolor[HTML]{C0C0C0}71.1 & 44.2 \\
 & \textit{no synthetic} & \cellcolor[HTML]{C0C0C0}76.4 & \cellcolor[HTML]{C0C0C0}\textbf{74.8} & \textbf{60.4} \\ \hline
\multirow{2}{*}{\textbf{RoBERTa}} & Default & \cellcolor[HTML]{C0C0C0}\textbf{85.0} & \cellcolor[HTML]{C0C0C0}81.2 & 51.8 \\
 & \textit{no synthetic} & \cellcolor[HTML]{C0C0C0}83.5 & \cellcolor[HTML]{C0C0C0}\textbf{83.4} & \textbf{63.1} \\ \hline
\multirow{2}{*}{\textbf{SpanBERT}} & Default & \cellcolor[HTML]{C0C0C0}\textbf{86.0} & \cellcolor[HTML]{C0C0C0}76.0 & 46.0 \\
 & \textit{no synthetic} & \cellcolor[HTML]{C0C0C0}82.2 & \cellcolor[HTML]{C0C0C0}\textbf{80.8} & \textbf{61.5} \\ \hline
\multirow{2}{*}{\textbf{Average}} & Default & \cellcolor[HTML]{C0C0C0}\textbf{83.3} & \cellcolor[HTML]{C0C0C0}76.1 & 47.3 \\
 & \textit{no synthetic} & \cellcolor[HTML]{C0C0C0}80.7 & \cellcolor[HTML]{C0C0C0}\textbf{79.7} & \textbf{61.7} \\ \hline
\end{tabular}
}
\caption{Performance of models fine-tuned on SQuAD 2.0 using Default training method and our proposed training method but without augmented synthetic answers, each averaged over five runs with random initialization. The performance on in-domain samples are highlighted in gray cells.}
\label{tab:squad2-nomultians-ood}
\end{table}

%% file: tables/agent-ood.tex
\begin{table}[ht]
\centering
\resizebox{\linewidth}{!}{
\begin{tabular}{lc|ccc}
\hline
\multicolumn{2}{c|}{\multirow{2}{*}{\begin{tabular}[c]{@{}c@{}}\textbf{\textit{Train Set:}} \\ \textbf{\textit{SQuAD AGent}} \end{tabular}}} & \multicolumn{3}{c}{\textbf{SQuAD}} \\ 
 &  & \cellcolor[HTML]{C0C0C0}has-ans & no-ans & \cellcolor[HTML]{C0C0C0}\textit{AGent} \\ \hline
\multirow{2}{*}{\textbf{BERT}} & Default & \cellcolor[HTML]{C0C0C0}\textbf{83.7} & 23.4 & \cellcolor[HTML]{C0C0C0}75.6 \\
 & \textit{\textbf{Ours}} & \cellcolor[HTML]{C0C0C0}80.3 & \textbf{30.1} & \cellcolor[HTML]{C0C0C0}\textbf{81.2} \\ \hline
\multirow{2}{*}{\textbf{RoBERTa}} & Default & \cellcolor[HTML]{C0C0C0}\textbf{87.7} & 30.2 & \cellcolor[HTML]{C0C0C0}84.4 \\
 & \textit{\textbf{Ours}} & \cellcolor[HTML]{C0C0C0}85.7 & \textbf{35.7} & \cellcolor[HTML]{C0C0C0}\textbf{88.8} \\ \hline
\multirow{2}{*}{\textbf{SpanBERT}} & Default & \cellcolor[HTML]{C0C0C0}\textbf{87.3} & 28.6 & \cellcolor[HTML]{C0C0C0}76.5 \\
 & \textit{\textbf{Ours}} & \cellcolor[HTML]{C0C0C0}83.6 & \textbf{36.6} & \cellcolor[HTML]{C0C0C0}\textbf{86.0} \\ \hline
\multirow{2}{*}{\textbf{Average}} & Default & \cellcolor[HTML]{C0C0C0}\textbf{86.2} & 27.4 & \cellcolor[HTML]{C0C0C0}78.8 \\
 & \textit{\textbf{Ours}} & \cellcolor[HTML]{C0C0C0}83.2 & \textbf{34.1} & \cellcolor[HTML]{C0C0C0}\textbf{85.3} \\ \hline
\end{tabular}
}
\caption{Performance of models fine-tuned on SQuAD \textit{AGent} using Default training method and our proposed training method, each averaged over five runs with random initialization. The performance on in-domain samples are highlighted in gray cells.}
\label{tab:agent-ood}
\end{table}

%% file: tables/agent-robustness.tex
\begin{table}[ht]
\centering
\resizebox{\linewidth}{!}{
\begin{tabular}{lc|c|cc|c}
\hline
\multicolumn{2}{c|}{\multirow{2}{*}{\begin{tabular}[c]{@{}c@{}}\textbf{\textit{Train Set:}} \\ \textbf{\textit{SQuAD AGent}} \end{tabular}}} & \multicolumn{1}{l|}{\multirow{2}{*}{Orig}} & \multicolumn{2}{c|}{\textbf{\begin{tabular}[c]{@{}c@{}}Adversarial \\ Attack\end{tabular}}} & \multicolumn{1}{l}{\multirow{2}{*}{$\Delta \downarrow$}} \\ \cline{4-5}
\multicolumn{2}{l|}{} & \multicolumn{1}{l|}{} & \textit{AOS} & Negation & \multicolumn{1}{l}{} \\ \hline
\multirow{2}{*}{\textbf{BERT}} & Default & \textbf{83.7} & 61.0 & 44.5 & 30.7 \\
 & \textit{\textbf{Ours}} & 80.3 & \textbf{67.0} & \textbf{57.1} & \textbf{18.3} \\ \hline
\multirow{2}{*}{\textbf{RoBERTa}} & Default & \textbf{87.7} & 68.6 & 46.4 & 30.2 \\
 & \textit{\textbf{Ours}} & 85.7 & \textbf{75.4} & \textbf{64.4} & \textbf{15.8} \\ \hline
\multirow{2}{*}{\textbf{SpanBERT}} & Default & \textbf{87.3} & 66.8 & 37.4 & 35.2 \\
 & \textit{\textbf{Ours}} & 83.6 & \textbf{72.2} & \textbf{65.9} & \textbf{14.6} \\ \hline
\multirow{2}{*}{\textit{\textbf{Average}}} & Default & \textbf{86.2} & 65.5 & 42.8 & 30.0 \\
 & \textit{\textbf{Ours}} & 83.2 & \textbf{71.5} & \textbf{62.5} & \textbf{16.2} \\ \hline
\end{tabular}
}
\caption{Robustness of models fine-tuned on SQuAD \textit{AGent} using Default training method and our proposed training method.}
\label{tab:agent-attack-performance}
\end{table}

%% file: tables/squad2-nomultians-robustness.tex
\begin{table}[ht]
\centering
\resizebox{\linewidth}{!}{
\begin{tabular}{lc|c|cc|c}
\hline
\multicolumn{2}{c|}{\multirow{2}{*}{\begin{tabular}[c]{@{}c@{}}\textbf{\textit{Train Set:}} \\ \textbf{\textit{SQuAD 2.0}} \end{tabular}}} & \multicolumn{1}{l|}{\multirow{2}{*}{Orig}} & \multicolumn{2}{c|}{\textbf{\begin{tabular}[c]{@{}c@{}}Adversarial \\ Attack\end{tabular}}} & \multicolumn{1}{l}{\multirow{2}{*}{$\Delta \downarrow$}} \\ \cline{4-5}
\multicolumn{2}{l|}{} & \multicolumn{1}{l|}{} & \textit{AOS} & Negation & \multicolumn{1}{l}{} \\ \hline 
\multirow{2}{*}{\textbf{BERT}} & Default & \textbf{78.8} & \textbf{52.2} & \textbf{27.5} & 38.9 \\
 & \textit{no synthetic} & 76.4 & 49.6 & 26.3 & \textbf{38.4} \\ \hline
\multirow{2}{*}{\textbf{RoBERTa}} & Default & \textbf{85.0} & \textbf{56.1} & \textbf{30.9} & 41.5 \\
 & \textit{no synthetic} & 83.5 & 55.0 & 30.1 & \textbf{40.9} \\ \hline
\multirow{2}{*}{\textbf{SpanBERT}} & Default & \textbf{86.0} & \textbf{57.9} & \textbf{30.7} & \textbf{41.7} \\
 & \textit{no synthetic} & 82.2 & 53.0 & 22.5 & 44.4 \\ \hline
\multirow{2}{*}{\textit{\textbf{Average}}} & Default & \textbf{83.3} & \textbf{55.4} & \textbf{29.7} & \textbf{40.7} \\
 & \textit{no synthetic} & 80.7 & 52.5 & 26.3 & 41.3 \\ \hline
\end{tabular}
}
\caption{Robustness of models fine-tuned on SQuAD 2.0 using Default training method and our proposed training method but without augmented synthetic answers.}
\label{tab:squad2-nomultians-attack-performance}
\end{table}

%% file: sections/8-conclusion.tex
\section{Conclusion}
In this paper, we introduce a novel training methodology for EQA models aimed at enhancing their robustness against distribution shifts and adversarial attacks. Our new training method is characterized by a novel training loss for the EQA problem, as well as challenging the single-answer assumption by creating  a new “synthetic” answer span in a number of answerable questions. Our experimental findings demonstrate that models trained using our approach exhibit significant improvement on out-of-domain testing datasets. Furthermore, the robustness of these models against two tested types adversarial attacks is also significantly better than that of the default models. 

In Section \ref{sec:discussion}, we study the independent contributions of our new loss function and the augmented “synthetic” answers to the robustness of our models. Our analysis reveals that the new loss function specifically benefits the performance on information-seeking unanswerable questions. This improved performance of information-seeking unanswerable questions contribute to the robustness against distribution shifts of models trained on SQuAD 2.0 with our method. 

On the other hand, our training method challenges the single-answer assumption of many existing EQA datasets by creating ``synthetic'' answers for a number of answerable questions. Our experiments indicate that these ``synthetic'' answers significantly contribute to the robustness of models trained with our method on both SQuAD 2.0 and SQuAD \textit{AGent} against adversarial attacks. This finding strongly corroborates our initial hypothesis, suggesting that the longstanding single-answer assumption of many EQA training datasets is a learning shortcut for models that can significantly compromise their robustness. We believe this work highlights the importance of future Question Answering datasets that incorporate the possibility of multiple, non-contiguous answer spans, similar to the MultiSpanQA dataset \cite{li-etal-2022-multispanqa}.

\section*{Limitations}
We acknowledge certain limitations in our work. Our study primarily focuses on evaluating the proposed training methodology using multiple pre-trained transformers-based models in English. This does not guarantee that our method will maintain its effectiveness when applied to other languages.

\section*{Acknowledgements}
We would like to thank Gia-Huy Do for his insightful discussions on the early versions of our proposed loss function and the derivation of unanswerable sequences (see Appendix \ref{appendix:lamba-dis}), and Anthony Silveira for his support with technical issues during our experiments.
We would like to express our gratitude to The William G. and Mary Ellen Bowen Research Endowment, The Laurie and David Hodgson Faculty Support Endowment, and the Denison University Research Fund for their generous support of this research.

%% file: sections/appendix.tex
\input{tables/dummy}

\appendix
\section{Derivation on Unanswerable Sequence}
\label{appendix:lamba-dis}

Let us consider the $k^{th}$ token in an \textbf{\textit{unanswerable}} sequence. Our objective is to ensure that the logit $s_k$  generally decreases if $s_k\geq 0$ after each training batch. To achieve this, we need the partial derivative of $L_{Ours}$ with respect to the start score $s_k$ of the $k^{th}$ token, i.e. $\frac{\lambda_{Tag} \partial L_{Tag}}{\partial s_k} + \frac{\lambda_{QA}\partial L_{QA}}{\partial s_k}$, remains positive whenever $s_k \geq 0$.

It is established that the partial derivative of the tagging loss $L_{Tag}$ with respect to the score $s_k$, $\frac{\partial L_{Tag}}{\partial s_k}$, is positive. Nonetheless, there is no assurance that the partial derivative of the question-answering loss $L_{QA}$ with respect to $s_k$, $\frac{\partial L_{QA}}{\partial s_k}$, will also be positive.

Firstly, we assume that both Tagging weight $\lambda_{Tag}$ and Question Answering weight $\lambda_{QA}$ are positive. We then have that
\begin{align*} 
\lambda_{Tag}&\frac{\partial L_{Tag}}{\partial s_k}\\
&= -\lambda_{Tag}\frac{d}{ds_k}\left[ \log(1- \frac{1}{1 + \exp(-s_k)})\right] \\
&= -\lambda_{Tag} \frac{\frac{d}{ds_k}\left[1- \frac{1}{1 + \exp(-s_k)}\right]}{1- \frac{1}{1 + \exp(-s_k)}}\\
&= -\lambda_{Tag} \frac{\frac{d}{ds_k}[1 + \exp(-s_k)]}{(1 + \exp(-s_k))^2 (1- \frac{1}{1 + \exp(-s_k)})}\\
&= \lambda_{Tag} \frac{\exp(-s_k)}{(1 + \exp(-s_k))^2 - (1 + \exp(-s_k))}\\
&= \lambda_{Tag} \frac{1}{1 + \exp(-s_k)}= \lambda_{Tag}(\frac{\exp(s_k)}{1 + \exp(s_k)})
\end{align*}
\begin{align*}
\lambda_{QA}&\frac{\partial L_{QA}}{\partial s_k} \\&= \lambda_{QA}\frac{\partial}{\partial s_k}\left[ -\Sigma_{k=1}^n \log \frac{\exp(s_k)}{\Sigma_{i=1}^n \exp(s_i)}y^s_k \right]\\
&= \lambda_{QA}\frac{\partial}{\partial s_k}\left[ -\Sigma_{k=1}^n \log \frac{\exp(s_k)}{\Sigma_{i=1}^n \exp(s_i)}\frac{1}{n} \right]\\
&= \frac{\lambda_{QA}}{n}\biggl(\frac{(n-1)\exp(s_k)}{\Sigma_{i=1}^n \exp(s_i)} \\& \qquad \qquad- \frac{\Sigma_{i=1}^n \exp(s_i) - \exp(s_k)}{\Sigma_{i=1}^n \exp(s_i)} \biggr)\\
&= \frac{\lambda_{QA}}{n}\left(\frac{n\exp(s_k)}{\Sigma_{i=1}^n \exp(s_i)} - 1 \right)\\
&=\lambda_{QA} \left(-\frac{1}{n} + \frac{\exp(s_k)}{\Sigma_{i=1}^n \exp(s_i)}\right)> -\frac{\lambda_{QA}}{n}
\end{align*}
Because $s_k \geq 0$, we know that $\frac{\exp(s_k)}{1 + \exp(s_k)} \geq \frac{1}{2}$. Therefore, we can derive that
\begin{align*}
    \lambda_{Tag}\frac{\partial L_{Tag}}{\partial s_k} &+  \lambda_{QA}\frac{\partial L_{QA}}{\partial s_k}\\ &>  \lambda_{Tag}(\frac{\exp(s_k)}{1 + \exp(s_k)}) -\frac{\lambda_{QA}}{n} \\&\geq \frac{\lambda_{Tag}}{2} - \frac{\lambda_{QA}}{n}
\end{align*}

Consequently, the partial derivative of the overall loss ($L_{Ours}$) with respect to the score $s_k$, $\frac{\partial L_{Ours}}{\partial s_k}$, will be positive whenever $s_k \geq 0$ if the ratio of $\frac{\lambda_{Tag}}{\lambda_{QA}} > \frac{2}{n}$. In our experiments, the number of tokens in a question-context sequence is set to $n=384$. We set $\lambda_{Tag} = 1$ and $\lambda_{QA} = 2$. Therefore, $\frac{\lambda_{Tag}}{\lambda_{QA}} = \frac{1}{2}> \frac{2}{384}$.

\section{Synthetic Answers}
\label{appendix:synthetic}
\subsection{Generate Synthetic Answers}
Table \ref{tab:synthetic-example} illustrates the incorporation of Synthetic answers into the context of $20\%$ of the answerable questions within the training set, serving as an example of our augmentation approach.

Incorporating ``synthetic'' answers into contexts of answerable questions involves three steps:
\begin{enumerate}
    \item Creating fake answers that differ from the ground truth answers annotated by human crowdsource workers.
        \begin{enumerate}
        \item We re-match each answerable question with 10 new contexts.
        \item We train 10 models on SQuAD 2.0 and obtain their predictions on the re-matched question-context pairs.
        \item For each answerable question, we extract the answer span that is most frequently predicted by the models.
        \end{enumerate}
    In this step, we ensure that the extracted spans are different from the corresponding ground truth answers, with F1 score lower than $0.2$. Through this method, we can extract relevant and plausible answers that can serve as ``synthetic'' answers for the corresponding questions.
    \item Given the fake answer and the original question, we use ChatGPT-turbo3.5 to convert them into a natural statement. We use the prompt:
    {\ttfamily
    \begin{align*}
    & \text{Given the question and its answer,}\\
    & \text{write a statement:} \\
    & \text{Example:} \\
    & \quad\text{<example1>} \\
    & \quad\text{<example2>} \\
    & \text{Question: <question>} \\
    & \text{Answer: <answer>} \\
    & \text{Statement: ...}
    \end{align*}
    }

    \item We then insert the newly created statement into the original context at a random position between existing sentences. We utilize SpaCy's pipeline \footnote{https://github.com/explosion/spaCy} to perform sentence boundary detection on original contexts. 
\end{enumerate}
\subsection{Do Synthetic Answers Cause Misleading Information?}

While generating ``synthetic'' answers for training our proposed models, we intentionally condition the generated answers to differ from the ground truth. As a result, these synthetic answers are factually incorrect. Consequently, training EQA models on these synthetic answers may lead to issues that model may extract biased or misleading information during the testing phase.

In this section, we investigate the risk of misleading information when training our models with ``synthetic'' answers. In this experiment, we use RoBERTa models trained with our proposed training approach, which includes a new training loss function and synthetic answers. We define an EQA model as not extracting misleading information if it refrains from extracting the synthetic answers when the provided context lacks sufficient information to support them.

To test whether synthetic answers induce misleading information, we evaluate our models on a modified version of the training set. 
For each training sample $(q,c',a')$, where $c'$ contains a sentence with a synthetic answer, we replace that sentence with only the synthetic answer. For example, we modify ``In 1948, the UN General Assembly adopted the Convention on the Prevention and Punishment of the Crime of Genocide (CPPCG) which defined the crime of genocide for the first time.'' to ``Convention on the Prevention and Punishment of the Crime of Genocide (CPPCG)''. In this scenario, lacking information about CPPCG renders it no longer an answer. We then compare the model's performance on these modified training samples with its performance on the corresponding $(q,c,a)$ samples from the original SQuAD 2.0 dataset (unmodified and no synthetic answer). 

The results indicate no significant difference as the F1 score drop from $78.8$ to $78.4$. This difference is not statistically significant, and the decline can largely be attributed to errors in determining the start and end bounds of the answers. Therefore, we conclude that when the context does not contain incorrect information supporting the "synthetic" answers, our models are likely to refrain from extracting them, thus avoiding misleading the users.

\subsection{Synthetic Answers in SQuAD 1.1}
\input{tables/v1-robustness}
The single-answer assumption is prevalent in many EQA datasets, both with and without unanswerable questions. In this section, we evaluate the effectiveness of our proposed "synthetic" answers on SQuAD 1.1, an EQA dataset without unanswerable questions, providing a comprehensive analysis of the impact of the single-answer assumption.

We train models on SQuAD 1.1 using our proposed methodology and with ``synthetic'' answers without the sequence tagging loss. We then compare these models (referred to as ``\textbf{Ours}'' and``\textit{NoTagging}'') with those trained using the default loss function, also trained on SQuAD 1.1. This experiment is designed to study the contributions of the ``synthetic'' answers to the robustness of our models in the EQA settings with answerable questions only.

The results in Table \ref{tab:v1-robustness} demonstrate that our models maintain robustness against adversarial attacks even in settings without unanswerable questions. Additionally, although the Sequence Tagging loss was designed for scenarios with unanswerable questions, it does not significantly affect the performance or robustness of EQA models in settings where all questions are answerable.

\section{Details for Models Training}
The input of a question-context pair into the pre-trained model is in the form of \texttt{[CLS]}\textit{<Question>}\texttt{[SEP]}\textit{<Context>}, with \texttt{[CLS]} and \texttt{[SEP]} as special tokens of pre-trained tokenizer accompanying the pre-trained model. After getting embeddings for each token, we feed its final embedding into a start and end token classifiers.

We train all models with batch size of $8$ for $3$ epochs. The maximum sequence length is set to 384 tokens. We use the AdamW optimizer \cite{loshchilov2018decoupled} with an initial learning rate of $2 \cdot 10^{-5}$, and $\beta_1 = 0.9$, $\beta_2 = 0.999$. We use a single NVIDIA GeForce RTX 3080 for training and evaluating models. The training process for $3$ epochs takes approximately $150$ minutes.

%% file: tables/dummy.tex
\begin{table*}[ht]
\centering
\resizebox{15cm}{!}{%
\begin{tabular}{clll}
\hline
Types & Question & Attacked Context & \begin{tabular}[c]{@{}c@{}}Ground Truth \\ Answer \end{tabular}\\ \hline
Original  & \begin{tabular}[c]{@{}l@{}}In 1948, what general\\ assembly resolution\\ established genocide\\ as a prosecutable act?\end{tabular}& \begin{tabular}[c]{@{}l@{}}[...] Lemkin successfully campaigned for the\\ universal acceptance of international laws\\ defining and forbidding genocides. In 1948, \\the UN General Assembly adopted the\\ \textcolor{blue}{\textbf{\textit{Convention on the Prevention and}}} \\\textcolor{blue}{\textbf{\textit{Punishment of the Crime of Genocide }}}\\\textcolor{blue}{\textbf{\textit{(CPPCG)}}} which defined the crime of\\ genocide for the first time. [...]\end{tabular} & \begin{tabular}[c]{@{}l@{}}\textcolor{blue}{\textbf{\textit{Convention on the Prevention}}} \\\textcolor{blue}{\textbf{\textit{and Punishment of the Crime of}}}\\\textcolor{blue}{\textbf{\textit{Genocide (CPPCG)}}}\end{tabular} \\ \hline

\begin{tabular}[c]{@{}c@{}}With \\``synthetic''\\ answer\end{tabular}  & \begin{tabular}[c]{@{}l@{}}In 1948, what general\\ assembly resolution\\ established genocide\\ as a prosecutable act?\end{tabular}& \begin{tabular}[c]{@{}l@{}}[...] Lemkin successfully campaigned for the\\ universal acceptance of international laws\\ defining and forbidding genocides.  \textcolor{red}{\textbf{In 1948,}}\\  \textcolor{red}{\textbf{Resolution 46/3 established genocide as a}} \\  \textcolor{red}{\textbf{prosecutable act}}. In 1948, the UN General\\ Assembly adopted the \textcolor{blue}{\textbf{\textit{Convention on the}}} \\\textcolor{blue}{\textbf{\textit{Prevention and Punishment of the Crime of}}}\\\textcolor{blue}{\textbf{\textit{Genocide (CPPCG)}}} which defined the crime\\ of genocide for the first time. [...]\end{tabular} & \begin{tabular}[c]{@{}l@{}}\textcolor{blue}{\textbf{\textit{Convention on the Prevention}}} \\\textcolor{blue}{\textbf{\textit{and Punishment of the Crime of}}}\\\textcolor{blue}{\textbf{\textit{Genocide (CPPCG)}}} \\ \\ \textcolor{red}{\textbf{Resolution 46/3}}\end{tabular} \\ \hline
\end{tabular}
}
\caption{An example of ``synthetic'' answers.}
\label{tab:synthetic-example}
\end{table*}

%% file: tables/v1-robustness.tex
\begin{table}[ht]
\resizebox{\linewidth}{!}{

\begin{tabular}{lc|c|c|c}
\hline
\multicolumn{2}{c|}{\multirow{2}{*}{}} & \multirow{2}{*}{Original} & \textbf{\begin{tabular}[c]{@{}c@{}}Adversarial \\ Attack\end{tabular}} & \multirow{2}{*}{$\Delta \downarrow$} \\ \cline{4-4}
\multicolumn{2}{c|}{} &  & \textit{AOS} &  \\ \hline
\multirow{3}{*}{\textbf{BERT}} & Default & \textbf{88.2} & 62.7 & 25.5 \\
 & \textit{NoTagging} & 87.8 & \textbf{70.5} & \textbf{17.3} \\
 & \textbf{Ours} & 87.7 & 69.1 & 18.6 \\ \hline
\multirow{3}{*}{\textbf{RoBERTa}} & Default & \textbf{92.1} & 70.2 & 21.9 \\
 & \textit{NoTagging} & 91.9 & \textbf{75.9} & \textbf{16.0} \\
 & \textbf{Ours} & 91.8 & 75.7 & 16.1 \\ \hline
\multirow{3}{*}{\textbf{SpanBERT}} & Default & \textbf{91.3} & 67.2 & 24.1 \\
 & \textit{NoTagging} & 91.2 & \textbf{75.1} & \textbf{16.1} \\
 & \textbf{Ours} & 90.6 & 73.8 & 16.9 \\ \hline
\multirow{3}{*}{\textbf{Average}} & Default & \textbf{90.5} & 66.7 & 23.8 \\
 & \textit{NoTagging} & 90.3 & \textbf{73.8} & \textbf{16.5} \\
 & \textbf{Ours} & 90.0 & 72.9 & 17.1 \\ \hline
\end{tabular}
}
\caption{Robustness against adversarial attacks of models fine-tuned on SQuAD 1.1 using Default training method and our proposed training method. The table also includes an ablation study on our proposed training method without Sequence Tagging Loss.}
\label{tab:v1-robustness}
\end{table}